\documentclass[conference]{IEEEtran}
\IEEEoverridecommandlockouts
\usepackage{cite}
\usepackage{amsmath,amssymb,amsfonts}
\usepackage{algorithmic}
\usepackage{graphicx}
\usepackage{subfigure}
\usepackage{textcomp}
\usepackage{xcolor}
\usepackage{url}
\usepackage{multirow}
\usepackage{booktabs}
\usepackage{graphicx} 
\usepackage{algorithmic}
\usepackage{algorithm}

\def\BibTeX{{\rm B\kern-.05em{\sc i\kern-.025em b}\kern-.08em
    T\kern-.1667em\lower.7ex\hbox{E}\kern-.125emX}}

\usepackage{listings}
\definecolor{codegreen}{rgb}{0,0.6,0}
\definecolor{codegray}{rgb}{0.5,0.5,0.5}
\definecolor{codepurple}{rgb}{0.58,0,0.82}
\definecolor{backcolour}{rgb}{0.95,0.95,0.92}
\definecolor{gray1}{HTML}{E6E6E6}

\makeatletter
\newcommand\footnoteref[1]{\protected@xdef\@thefnmark{\ref{#1}}\@footnotemark}
\makeatother
\lstdefinestyle{myStyle}{
    backgroundcolor=\color{gray1},   
    commentstyle=\color{codegreen},
    stringstyle=\color{codepurple},
    basicstyle=\ttfamily\footnotesize,
    breakindent=0em,
    breakatwhitespace=true,         
    breaklines=true,        
    columns=flexible,
    keepspaces=true,                 
    numbers=left,       
    numbersep=5pt,                  
    showspaces=false,                
    showstringspaces=false,
    showtabs=false,                  
    tabsize=2,
}
\IEEEoverridecommandlockouts
\IEEEpubid{\makebox[\columnwidth]{ 979-8-3503-5067-8/24/\$31.00~\copyright2024 IEEE \hfill} 
\hspace{\columnsep}\makebox[\columnwidth]{ }}
\begin{document}

\title{3D Building Generation in Minecraft via Large Language Models
}

\author{\IEEEauthorblockN{Shiying Hu, Zengrong Huang, Chengpeng Hu and Jialin Liu}
\IEEEauthorblockA{\textit{Guangdong Key Laboratory of Brain-inspired Intelligent Computation, 
Department of Computer Science and Engineering,} \\ \textit{Southern University of Science and Technology, Shenzhen, China}
}

\thanks{S. Hu and Z. Huang contributed equally to this work.}
\thanks{This paper has been accepted by IEEE Conference on Games.}
}

\maketitle
\IEEEpubidadjcol
\begin{abstract}
Recently, procedural content generation has exhibited considerable advancements in the domain of 2D game level generation such as \textit{Super Mario Bros.} and \textit{Sokoban} through large language models (LLMs). To further validate the capabilities of LLMs, this paper explores how LLMs contribute to the generation of 3D buildings in a sandbox game, \textit{Minecraft}. 
We propose a \textit{Text to Building in Minecraft (T2BM)} model, which involves refining prompts, decoding interlayer representation and repairing. Facade, indoor scene and functional blocks like doors are supported in the generation.
Experiments are conducted to evaluate the completeness and satisfaction of buildings generated via LLMs. 
It shows that LLMs hold significant potential for 3D building generation. Given appropriate prompts, LLMs can generate correct buildings in \textit{Minecraft} with complete structures and incorporate specific building blocks such as windows and beds, meeting the specified requirements of human users.
\end{abstract}


\section{Introduction}
Procedural content generation (PCG) involves automatically generating specific content such as environments and levels in games~\cite{shaker2016procedural,summerville2018procedural,liu2021deep,guzdial2022procedural}.
Recent emergence of LLMs~\cite{openai2023chatgpt,touvron2023llama,zhao2023survey} have inspired various applications in PCG~\cite{gallotta2024large}.
The capabilities of LLMs are crucial in enabling PCG to fulfil complex requirements from human feedback. For instance, Todd \textit{et al.}~\cite{todd2023level} fine-tuned LLMs to generate novel and playable \emph{Sokoban} levels. Sudhakaran \textit{et al.}~\cite{shyam2024mariogpt} created MarioGPT to generate \emph{Super Mario Bros.}'s levels. Instead of using tokenizer~\cite{todd2023level}, MarioGPT enables users to directly write prompts, e.g., ``\textit{some pipes, little enemies}" to obtain expected levels~\cite{shyam2024mariogpt}. 
Nasir \textit{et al.}~\cite{nasir2023practical} constructed a two-stage workflow to fine-tune GPT-3, in which levels generated by both humans and LLMs are involved for training. 
Besides, an LLM-based competition called ChatGPT4PCG was held in 2023, aiming at generating \textit{Angry Birds} levels via LLMs~\cite{Abdullah2024ChatGPT4PCG}. 
However, all the aforementioned works focused on 2D level generation. 
Applying LLMs to 3D game content generation is rarely explored.

Considering a 3D environment, the additional dimension (e.g., height) necessitates the consideration of extra attributes and a higher-dimensional representation. \textit{Minecraft} is a suitable testbed~\cite{hu2024games} for 3D building generation for global popularity, well-defined grid structure and high editability.  Generative design in \textit{Minecraft} competition (GDMC) encourages the application of PCG in 3D building generation~\cite{salge2018generative}. 
Traditional JSON lookup methods like Nys \textit{et al.}~\cite{nys2020cityjson} involved extracting keywords from user inputs and matching them to predefined JSON templates. However, it often fails to understand complex requests.
Green \textit{et al.}~\cite{green2019organic} combined constrained growth and cellular automata method for generating buildings with ASCII representation.
Merino~\textit{et al.}~\cite{merino2023interactive} proposed an interactive evolutionary algorithm, in which users choose preferred buildings. However, it takes time to obtain a suitable building.
Huang \textit{et al.}~\cite{huang2023generating} searched city layouts to place redstone-style buildings.
Barthet \textit{et al.}~\cite{barthet2022open} generated buildings by searching latent space and constructing rules.
Jiang \textit{et al.}~\cite{jiang2022learning} incorporated reinforcement learning to generate 3D levels.
Furthermore, Maren \textit{et al.}~\cite{awiszus2021world} introduced World-GAN to generate 3D scenarios such as deserts. Very recently, Earle \textit{et al.}~\cite{earle2024dreamcraft} trained quantised neural radiance fields to facilitate text-to-3D generation.
However, functional blocks like doors and indoor scene generation are not addressed~\cite{earle2024dreamcraft}.
Directly incorporating human feedback such as language in 3D building generation in \textit{Minecraft} remains an under-explored, non-trivial challenge.

 To investigate into the above topic, this paper proposes a \textit{Text to Building in Minecraft (T2BM)} model, which leverages capabilities of LLMs in 3D building generation considering facade, indoor scene and functional blocks such as doors and beds. 
T2BM accepts simple prompts from users as input and generates buildings encoded by an interlayer, which defines the transformation between text and digital content.
Based on T2BM, players or designers can construct buildings quickly without repeatedly placing blocks one by one, while the human-crafted prompt is not necessarily detailed.
Experiments with GPT-3.5 and GPT4 demonstrate that T2BM can generate complete buildings, while aligning with human instructions.

\begin{figure*}[htbp]
	\centering 
	\includegraphics[width=1\textwidth]{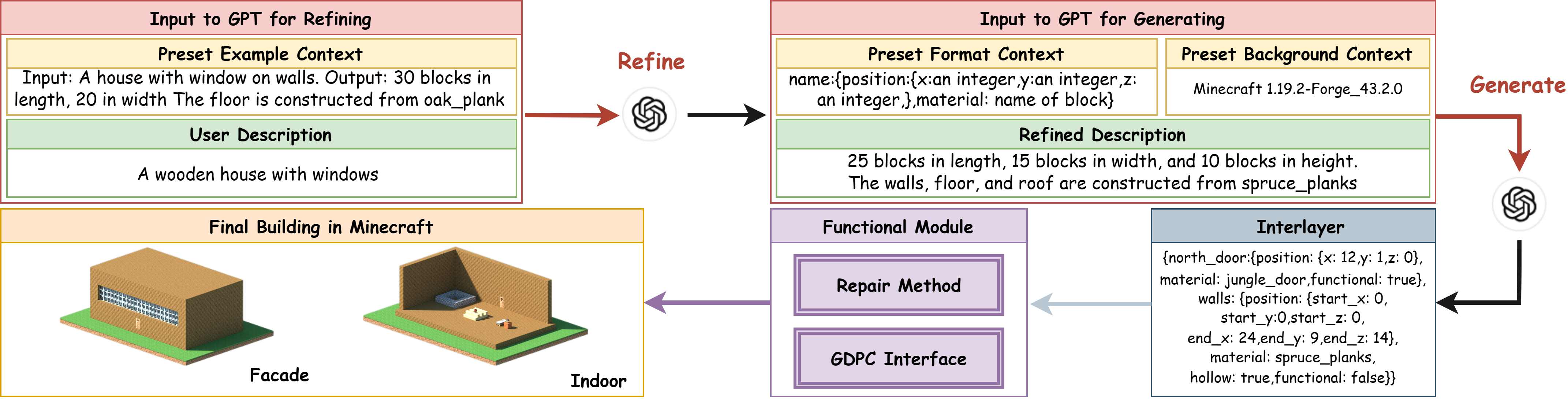}
	\caption{Workflow of \textit{Text to Building in Minecraft (T2BM)} model.}
	\label{fig_mom0}
\end{figure*}

\section{Text to Building in Minecraft (T2BM)}
As depicted in Fig.~\ref{fig_mom0}, T2BM receives a simple user input and outputs a complete building in \textit{Minecraft}. Initially, T2BM forwards the user's description along with contexts like some detailed building description examples to an LLM for refinement. Then, the refined prompt is combined with a format example and background to structure the final instruction, according to which an LLM produces a building encoded by the interlayer. Subsequently, discrepancies in this interlayer are corrected using a repairer. Finally, T2BM decodes the generated building in \textit{Minecraft}.
T2BM is composed of three core modules, namely input refining, interlayer, and repairing.

\subsection{Input refining module}
\label{sec:refine}
In T2BM, the prompt to LLMs comprises two main components: the description provided by the user and detailed building description examples. Due to the critical impact of the prompt's quality on the outcomes, it is essential for LLMs to first generate a refined prompt. This preliminary step enhances the overall quality of the resulting buildings. The \textit{refining} strategy is informed by the prompt engineering guide\footnote{\url{https://github.com/dair-ai/Prompt-Engineering-Guide}}, ensuring that the modifications to the building description significantly improve the final output. Below is an example of a refined prompt, given ``\textit{A wooden house with windows}''.
\begin{lstlisting}[frame=bt, numbers=none, captionpos=b, abovecaptionskip=0pt, belowcaptionskip=0pt]
The wooden house is a quaint, homely structure, measuring 25 blocks in length, 15 blocks in width, and 10 blocks in height. The walls, floor, and roof are constructed from spruce_planks, lending the house a traditional, rustic appeal. The walls are interspersed with windows crafted from glass_panes, allowing the sunlight to permeate the house and offering panoramic views of the nearby scenery.
\end{lstlisting}
\subsection{Interlayer between prompts and buildings}
We introduce an \textit{interlayer} to transform text to recognisable content. 
The interlayer is designed in JSON to represent buildings in \textit{Minecraft}. Since JSON is a lightweight data format that simplifies the representation of complex building structures, it aids LLMs in comprehending and generating precise details of the building, including block locations and type. Besides, the interlayer allows to generate external facade and indoor of buildings simultaneously.
Users can also adjust the interlayer before generating the final building, which simplifies the correction of errors and reduces time cost.

The interlayer decomposes a building into various components, including the roof, doors, and windows, classified as either \textit{structural} or \textit{functional} sections, contingent upon their ability to interact with the player. Structural sections, such as walls and windows, do not possess unique interactive states and thus cannot interact with the player. On the other hand, functional sections like doors and crafting tables, can interact with the player, e.g., doors can be opened or closed, and crafting tables allow players to craft items. 

Each section contains two basic properties: \textit{material} and \textit{location}. Material refers to the used material type. In terms of location, structural sections are characterised by start and end coordinates, which define the extent of their placement from one spatial point to another within the architectural layout. In contrast, functional sections, including doors and crafting tables, are identified by a single placement coordinate, pinpointing their exact location in the structure. In addition, structural sections include the \textit{hollow} attribute to indicate whether they are solid or hollow, while functional sections feature the \textit{state} property to specify their particular characteristics. A \textit{format example of interlayer} is shown below.
\begin{lstlisting}[frame=bt, numbers=none, captionpos=b, abovecaptionskip=5pt, belowcaptionskip=10pt, language=Python]
{"wall":{ # structural section
    "location":{
        "start_x":0,"start_y":0,"start_z":0,
        "end_x":8,"end_y":6,"end_z":6},
    "material":"oak_planks","hollow":true },
"door":{ # functional section
    "location":{"x": 4,"y": 3,"z": 3},
    "material":"oak_door",
    "state" : {"facing":"south","hinge":"left"}},}
\end{lstlisting}

To generate buildings encoded by the interlayer, the refined prompt is combined with a format example of interlayer and background context.
The background contains some descriptions and requirements such as the edition of \textit{Minecraft} as different editions may have different names of materials. Here is an example of \textit{background context}~\footnote{\textit{Minecraft} version: Minecraft 1.19.2-Forge\_43.2.0}.

\noindent\begin{minipage}{\linewidth}
\begin{lstlisting}[frame=bt, numbers=none, captionpos=b, abovecaptionskip=5pt, belowcaptionskip=10pt]
The buildings in Minecraft are represented as grid and each block is located in a specific coordinate.
Now we want to generate a building in MC but represent with JSon format.  Please use identifiers in Minecraft 1.19.2-Forge_43.2.0.
\end{lstlisting}
\end{minipage}

\subsection{Repairing module}
LLMs sometimes introduce errors when generating interlayers, such as using upper camel case instead of underscores, or omitting block colours like ``\textit{white\_bed}". Therefore, it is essential to perform thorough checks and validations.
Four common errors and their corresponding handling strategies are considered.
\textit{Incomplete name}: Prefixes that denote features like colour or material, may be missing. The repairer will automatically complete the name with a default value such as \texttt{white} or \texttt{oak}.
\textit{Disallowed property}: Properties like \texttt{occupied} of bed that cannot be set by \textit{Minecraft} APIs will be ignored.
\textit{Illegal material}: For certain inputs, LLMs may generate block names that are unavailable in a specific edition of \textit{Minecraft}. They will be replaced with \texttt{oak\_planks} automatically.
\textit{Wrong naming style}: LLMs may generate material names with incorrect style, such as \texttt{Red Bed}.

After repairing the building in interlayer representation, T2BM feeds this interlayer into \textit{Minecraft} via generative design python client (GDPC)\footnote{\url{https://github.com/avdstaaij/gdpc}}, which facilitates the sequential generation of each section in the interlayer.

\section{Experiment}
Given that our building representation relies on the interlayer, the primary focus of our experiment is to investigate the ability of LLMs to accurately generate this interlayer. During the experiment, we evaluate the impacts of various prompts and different LLMs on the generation process. Examples are provided in \textbf{Supplementary Material}\footnote{\url{https://github.com/SUSTechGameAI/Text-to-Building-in-Minecraft}}. 

\subsection{Evaluation criteria}
A regular building should be fully sealed except for the door and include all specified blocks in the prompt. We proposed Algorithm~\ref{alg:CVA} to assess the correctness, including the completeness and satisfaction of generated buildings. 
The flood-fill algorithm~\cite{barthet2022open} is integrated into Algorithm~\ref{alg:CVA} to determine if all blocks are directly connected to the main body of the building. Flood-fill starts from a random initial point, recursively traverses the blocks connected to the current block, and saves all the accessed blocks. The satisfaction is checked by recording if all the materials requested in the prompt are present. 
The completeness is checked if a block belongs to the main structure and is directly connected to any blocks that form \textit{Corner}, \textit{Edge} or \textit{Plane}.
Algorithm~\ref{alg:CVA} classifies blocks by their connecting structure and removes blocks that do not belong to the main structure. Blocks that lack surrounding blocks to form a connected main structure are also excluded. 

\begin{algorithm}[htbp]
    \caption{Correctness assessment algorithm}
    \label{alg:CVA}
    \begin{algorithmic}[1]
        \REQUIRE $\mathcal{E}$, the editor interact with \textit{Minecraft}; $p_0$, the start point of building; $\mathcal{L}$, a list of required materials in prompt
        \ENSURE material constraint satisfaction $S$ and completeness constraint satisfaction $C$
        \STATE $S \leftarrow \FALSE,C \leftarrow \FALSE$ 
        \STATE $\mathcal{M}\leftarrow\{\},\mathcal{P}\leftarrow\{\},\mathcal{Q}\leftarrow <p_0>$ \COMMENT{\textit{materials appeared, points in main structure, and queue of visiting points}}
        \WHILE{$\mathcal{Q}$ is not empty}
            \STATE $p\leftarrow$ element pop from $\mathcal{Q}$
            \STATE add $p$ to $\mathcal{P}$; add the material of block at $p$ by $\mathcal{E}$ to $\mathcal{M}$
            \FOR{each $p' \in$ adjacent points of $p$}
            \STATE add $p'$ to $\mathcal{Q}$ \textbf{if} $p'$ not visited
            \ENDFOR
        \ENDWHILE   
        \STATE $S\leftarrow$ all the materials in $\mathcal{L}$ are presented in $\mathcal{M}$
        \WHILE{$\mathcal{P}$ is not empty \& size changing}
            \STATE remove points not forming connect structure from $\mathcal{P}$
        \ENDWHILE
        \STATE $C\leftarrow$ $\mathcal{P}$ is not empty\COMMENT{\textit{exists a complete main structure}}
        \RETURN $S$ and $C$
     \end{algorithmic}
\end{algorithm}

\begin{figure*}[htbp]
    \centering
    \includegraphics[width=1\textwidth, angle=0]{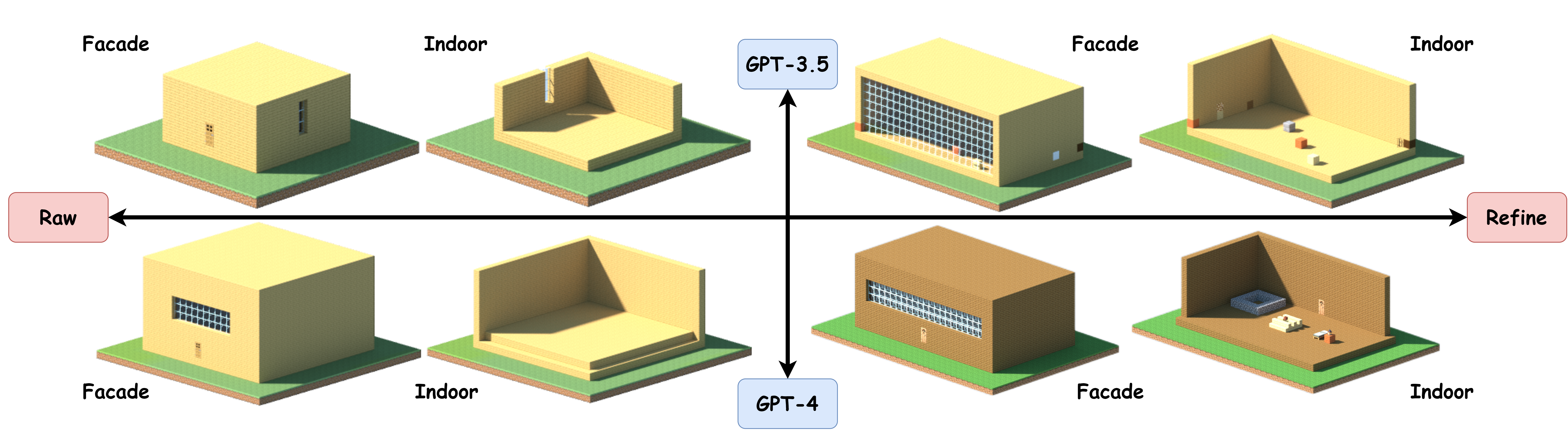}
    \caption{Buildings in the case of C$\wedge$S generated without and with refining given the user input ``\textit{A wooden house with windows}''.}
    \label{fig:allcs}
\end{figure*}

\subsection{Experimental setting and results}
GPT-3.5 and GPT-4 are chosen for their superior logical processing and the supportive OpenAI API that eases our experimental process. 
Two settings are considered to examine the contribution of refinement:
(i) raw: A simple raw user input is like ``\textit{A wooden house with windows}";
(ii) refined: A good refined description by LLMs after accepting user input (cf. Section~\ref{sec:refine}). Both are evaluated 50 times. Results of completeness and satisfaction constraints checking are shown in  Tab.~\ref{tab:cs}. ``C" denotes the ratio of complete outputs among all outputs. ``S" denotes the ratio of outputs that satisfy the material constraint, i.e., the user's input, among all outputs. ``$\neg$C$\wedge$$\neg$S" represents the ratio of outputs that meet neither ``C" nor ``S". ``C$\wedge$$\neg$S" and ``$\neg$C$\wedge$S" represents the ratio of outputs that only meet ``C" and ``S", respectively. ``C$\wedge$S" represents the ratio of outputs that meet both ``C" and ``S".

\begin{table}[h!]
\begin{center}
\caption{Completeness and Satisfaction of buildings generated by GPT-3.5 and GPT-4 given Raw and Refined Prompts}
\begin{tabular}{l|c|c|c|c@{\hspace{1.8mm}}c@{\hspace{1.8mm}}c@{\hspace{1.8mm}}c@{\hspace{1.8mm}}}
\toprule
\textbf{Model} & \textbf{Prompt} & \textbf{C} & \textbf{S} & \textbf{$\neg$C$\wedge$$\neg$S} & \textbf{C$\wedge$$\neg$S} & \textbf{$\neg$C$\wedge$S} & \textbf{C$\wedge$S} \\
\midrule
\multirow{2}{*}{GPT-3.5} & Raw & 0.66 & 0.08 & \underline{0.34} & 0.58 & 0.00 & 0.08 \\
                         & Refined & 0.80 & 0.28 & 0.20 & 0.58 & 0.00 & \textbf{0.22}  \\
\midrule
\multirow{2}{*}{GPT-4}   & Raw & 0.80 & 0.12& \underline{0.20}  & 0.68 & 0.00 & 0.12 \\
                         & Refined & 0.82 & 0.48 & 0.08 & 0.44 & 0.10 & \textbf{0.38} \\
\bottomrule
\end{tabular}
\label{tab:cs}
\end{center}
\end{table}

\subsection{Discussions}
As shown in Tab.~\ref{tab:cs}, after the refinement, both completeness and satisfaction have been improved. Moreover, GPT-4 performs better than GPT-3.5. 
Fig.~\ref{fig:allcs} shows buildings that met both completeness
and satisfaction constraints (C$\wedge$S) generated by GPT-3.5 and GPT-4 with raw and refined prompts.

\subsubsection{Impact of prompt refinement}
Tab.~\ref{tab:cs} illustrates that refining prompts enhances the outputs of both GPT-3.5 and GPT-4. The ratio of generated buildings that satisfy both constraints increases from 0.08 to 0.22 by GPT-3.5 and from 0.12 to 0.38 by GPT-4.
With the assistance of the refined prompts, LLMs are capable of identifying specific items within the generated buildings and their respective locations, significantly reducing the workload. Given a list of legal materials or detailed building generation process in the refined prompt, the outputs may be more effective than before.
\subsubsection{Diminishing returns of completeness}
Tab.~\ref{tab:cs} also implies that generating complete buildings is harder than using legal materials.
While the increase in ``C" from raw to refined prompts is apparent, it is more significant in GPT-3.5 than GPT-4. Given that 80\% of the buildings generated by GPT-4 with raw prompts are complete, the gain owing to refinement is less remarkable.
GPT-4 can generate relatively complete buildings even being given raw prompts. It suggests that as LLMs become more sophisticated, they may be able to better grasp subtleties and context.
GPT-3.5's performance shows greater improvement with refined prompts, highlighting its higher sensitivity to the quality of input, which necessitates more precise prompts to achieve better performance.

\section{Conclusion}
This paper explores the application of LLMs for 3D building generation considering facade, indoor and functional blocks.
A \textit{Text to Building in Minecraft (T2BM)} model is proposed. T2BM refines user inputs, encodes buildings by an interlayer and fixes errors via repairing methods. Based on T2BM, users can generate buildings by only inputting simple prompts.
Our experiments validate the completeness and satisfaction of buildings generated by T2BM.
The importance of refinement and performance gaps between LLMs are also investigated.
In the future, we will integrate repairing to prompt guidelines and expand T2BM to other game environments.

\bibliographystyle{IEEEtran}
\bibliography{main}

\renewcommand{\appendixname}{Supplementary Material}
\newpage
\onecolumn
\appendix
\setcounter{figure}{0}
\setcounter{table}{0}

The supplementary material details the process of T2BM and generated building examples. 
\subsection{Interconnected components for completeness check}

\begin{figure}[htbp]
    \centering
    \subfigure[Corner in structure]{
	\includegraphics[width=0.3\textwidth, angle=0]{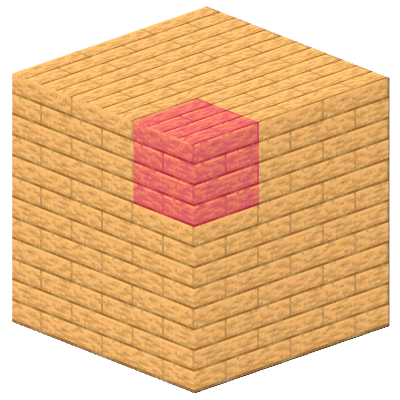}	
	\label{fig:corner}    
    }\hspace{1em}
    \subfigure[Edge in structure]{
	\includegraphics[width=0.3\textwidth, angle=0]{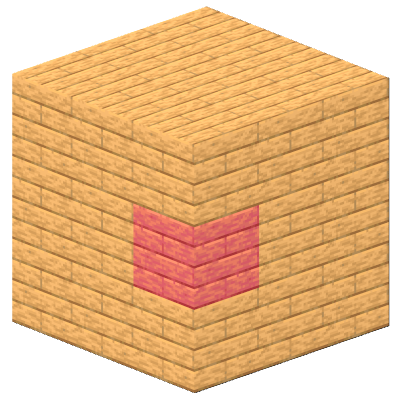}	
	\label{fig:edge}    
    }\hspace{1em}
    \subfigure[Plane in structure]{
	\includegraphics[width=0.3\textwidth, angle=0]{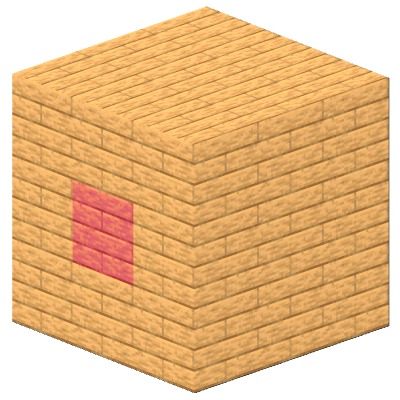}	
	\label{fig:plane}    
    }
    \caption{Three types of connecting structures.}
    \label{fig:three_type}
\end{figure}

The completeness is checked according to three interconnected components: 
\textit{Corner} (cf. Fig.~\ref{fig:corner}), \textit{Edge} (cf. Fig.~\ref{fig:edge}) and \textit{Plane} (cf. Fig.~\ref{fig:plane}). 
In Fig. \ref{fig:three_type}, the red block will be checked if it belongs to the main structure and is directly connected to any blocks that form one of the three connecting structures.

\subsection{Detailed prompts to LLMs in T2BM}

Below displays a user input and the preset context from which GPT-4 will generate the refined prompt. 

\begin{lstlisting}[frame=bt, numbers=none, captionpos=b, abovecaptionskip=5pt, belowcaptionskip=10pt]
    User Input = "A wooden house with windows"
    Context = 
    """
        Please help me expand this context.I will give you an example.You need a detailed description of the objects for the room.And give the material, and the specific orientation of the items.  The material should be identifiers in Minecraft 1.19.2-Forge_43.2.0!!!But please don't add"  minecraft:   Building blocks in minecraft can't have state facing or hinge "
    Example:
    "
        User Input: A wooden house with window on walls.
        Output: The wooden house is a charming, rustic structure, with dimensions of 30 blocks in length, 20 blocks in width, and 15 blocks in height. The walls, floor, and roof are made of oak_planks, giving the house a warm, cozy feel. The walls are unctuated with windows made of glass_panes, allowing natural light to flood into the house and offering views of the surrounding landscape.
        The house has a single, large room that serves multiple purposes. In the center of the room, there is a dining table made of spruce_planks, surrounded by four chairs made of the same material. The table is oriented east-west, with two chairs on each side. On the table, there is a flower_pot with a red_tulip, adding a touch of nature to the room.
        To the north of the room, there is a kitchen area with a crafting_table and a furnace, both made of stone. The crafting table is to the east of the furnace. There is also a chest made of birch_planks, filled with food items like bread and apples.
        On the south side of the room, there is a comfortable bed made of white_wool, with a bedside table made of jungle_planks. The bed is oriented north-south, with the head of the bed against the southern wall. The bedside table holds a lantern made of iron_ingot and torch, providing a soft light.
        The house has two doors, one on the east wall and one on the west wall. Both doors are made of dark_oak_planks and are oriented north-south. The eastern door leads to a small porch, while the western door opens to a lush garden.
        The wooden house is a perfect example of simple, rustic charm. Every block and item has been carefully chosen and placed to create a warm, inviting space that feels like home.
    "
    Output:
    """
\end{lstlisting}

~\\

Below displays the refined description generated by GPT-4, the preset format and the background. They make up the prompt, which will be passed to GPT-4 to generate the interlayer.
\begin{lstlisting}[frame=bt, numbers=none, captionpos=b, abovecaptionskip=5pt, belowcaptionskip=10pt]
    Refined Description: 
        The wooden house is a quaint, homely structure, spanning 25 blocks in length, 15 blocks in width, and 10 blocks in height. The walls, floor, and roof are constructed from spruce_planks, lending the house a traditional, rustic ambiance. The walls are interspersed with windows crafted from glass_panes, allowing the sunlight to permeate the house and offering panoramic views of the surrounding scenery.
The house comprises a single, spacious room that serves various functions. In the middle of the room, there is a coffee table made of birch_planks, encircled by three chairs made of the same material. The table is oriented north-south, with one chair on the north side and two on the south side. On the table, there is a flower_pot with a blue_orchid, adding a splash of color to the room.
To the west of the room, there is a kitchen area with a crafting_table and a furnace, both constructed from cobblestone. The crafting table is to the north of the furnace. There is also a chest made of oak_planks, stocked with food items like cooked_beef and carrots.
On the east side of the room, there is a cozy bed made of light_gray_wool, with a bedside table made of acacia_planks. The bed is oriented east-west, with the head of the bed against the eastern wall. The bedside table holds a lantern made of iron_ingot and torch, casting a gentle glow.
The house has two doors, one on the north wall and one on the south wall. Both doors are made of jungle_planks and are oriented east-west. The northern door leads to a small patio, while the southern door opens to a vibrant flower garden.
The wooden house is a beautiful representation of rustic simplicity. Each block and item has been meticulously selected and positioned to create a warm, welcoming space that feels like a true sanctuary.
    Format:
        """format of json:
        Please pay attention to the functional block,it should have only three attributes in position, which means it is a single block. And for a non-functional block, it should have six attributes. Building blocks in minecraft can't have state facing or hinge. For a two-square object such as a bed, you only need to give one coordinate.
        {   
        "name of building":{
            //there can be many parts in a building
            "name of part1 in this building":{
                "position":{
                    //this is an example of functional is false
                    "start_x": an integer,
                    "start_y": an integer,
                    "start_z": an integer,
                    "end_x": an integer,
                    "end_y": an integer,
                    "end_z": an integer,
                },
                "material" : "name of block in Minecraft",
                "hollow" : boolean value indicate whether this is a hollow part,
                "functional" : false
            }
            "name of part2 in this building":{
                "position":{
                    //this is an example of functional is true
                    "x": an integer,
                    "y": an integer,
                    "z": an integer,
                },
                "material" : "name of block in Minecraft",
                "hollow" : true
                "functional" : true
                "state": {
                    "facing": "south",
                    "hinge": "left"
                } some arguments for functional block
            }
        }   
        }
        In this format, all item compose a complete building, and for each item, we give some arguments like position to indicate where to place blocks, material to indicate which block to place, hollow to indicate hollow or not, functional to indicate functional or not, state(if functional) to indicate some state argument of block.
Background: 
        The buildings in Minecraft are represented as grid and each block is located in a specific coordinate. Now we want to generate a building in Minecraft but represent with JSON format. Please generate a realistic room, with various shapes, complex squares, not simple cuboids. Rooms should have doors, walls, ceilings and floors. Please use identifiers in Minecraft 1.19.2-Forge_43.2.0!!!
\end{lstlisting}
Below is the interlayer example. Based on this interlayer, buildings can be decoded in \textit{Minecraft} via the GDPC interface.
\begin{lstlisting}[frame=bt, numbers=none, captionpos=b, abovecaptionskip=5pt, belowcaptionskip=10pt]
{
    "wooden_house": {
        "walls": {
            "position": {
                "start_x": 0,
                "start_y": 0,
                "start_z": 0,
                "end_x": 24,
                "end_y": 9,
                "end_z": 14
            },
            "material": "spruce_planks",
            "hollow": true,
            "functional": false
        },
        "floor": {
            "position": {
                "start_x": 0,
                "start_y": 0,
                "start_z": 0,
                "end_x": 24,
                "end_y": 0,
                "end_z": 14
            },
            "material": "spruce_planks",
            "hollow": false,
            "functional": false
        },
        "roof": {
            "position": {
                "start_x": 0,
                "start_y": 10,
                "start_z": 0,
                "end_x": 24,
                "end_y": 10,
                "end_z": 14
            },
            "material": "spruce_planks",
            "hollow": false,
            "functional": false
        },
        "north_door": {
            "position": {
                "x": 12,
                "y": 1,
                "z": 0
            },
            "material": "jungle_door",
            "hollow": true,
            "functional": true,
            "state": {
                "facing": "south",
                "hinge": "left"
            }
        },
        "south_door": {
            "position": {
                "x": 12,
                "y": 1,
                "z": 14
            },
            "material": "jungle_door",
            "hollow": true,
            "functional": true,
            "state": {
                "facing": "north",
                "hinge": "right"
            }
        },
        "windows": {
            "position": {
                "start_x": 2,
                "start_y": 4,
                "start_z": 0,
                "end_x": 22,
                "end_y": 6,
                "end_z": 14
            },
            "material": "glass_pane",
            "hollow": true,
            "functional": false
        },
        "table": {
            "position": {
                "x": 12,
                "y": 1,
                "z": 7
            },
            "material": "birch_planks",
            "hollow": true,
            "functional": true
        },
        "chairs": {
            "position": {
                "start_x": 11,
                "start_y": 1,
                "start_z": 6,
                "end_x": 13,
                "end_y": 1,
                "end_z": 8
            },
            "material": "birch_stairs",
            "hollow": true,
            "functional": false
        },
        "flower_pot": {
            "position": {
                "x": 12,
                "y": 2,
                "z": 7
            },
            "material": "flower_pot",
            "hollow": true,
            "functional": true
        },
        "kitchen": {
            "position": {
                "start_x": 2,
                "start_y": 1,
                "start_z": 2,
                "end_x": 6,
                "end_y": 1,
                "end_z": 6
            },
            "material": "cobblestone",
            "hollow": true,
            "functional": false
        },
        "bed": {
            "position": {
                "x": 18,
                "y": 1,
                "z": 7
            },
            "material": "light_gray_bed",
            "hollow": true,
            "functional": true,
            "state": {
                "facing": "west",
                "part": "head"
            }
        },
        "bedside_table": {
            "position": {
                "x": 20,
                "y": 1,
                "z": 7
            },
            "material": "acacia_planks",
            "hollow": true,
            "functional": true
        },
        "lantern": {
            "position": {
                "x": 20,
                "y": 2,
                "z": 7
            },
            "material": "lantern",
            "hollow": true,
            "functional": true
        }
    }
}
\end{lstlisting}

~\\

This is the repairing method and GDPC interface. 
\begin{lstlisting}[frame=bt, numbers=none, captionpos=b, abovecaptionskip=5pt, belowcaptionskip=10pt]
            try:
                position = part['position']
            except Exception:
                whether = False
            if not whether:
                break
            if position.__contains__('x'):
                material = part['material']
                material = str(material).lower()
                material.replace(' ', '_')
                if material == 'door' or material == 'minecraft:door':
                    material = 'oak_door'
                elif material == 'bed' or material == 'minecraft:bed':
                    material = 'white_bed'
                elif material == 'iron_ingot' or material == 'canvas':
                    material = 'iron_block'
                state = ''
                if part.__contains__('state'):
                    state += '['
                    for s in part['state']:
                        if (str(part['state'][s]) == 'occupied' or
                                str(part['state'][s]) == 'open'):
                            continue
                        if material.__contains__(s):
                            state += s
                            state += '='
                            state += str(part['state'][s])
                            state += ','
                    state = state[0:len(state) - 1]
                    state += ']'
                try:
                    editor.placeBlock((x + position['x'], z + position['y'], y + position['z']),
                                  Block(material + state))
                except Exception as e:
                    whether = False
                if not whether:
                    break
            else:
                material = part['material']
                material = str(material).lower()
                material.replace(' ', '_')
                if material == 'planks' or material == 'minecraft:planks':
                    material = 'oak_planks'
                elif material == 'carpet' or material == 'minecraft:carpet':
                    material = 'white_carpet'
                elif material == 'stairs' or material == 'minecraft:stairs':
                    material = 'oak_stairs'
                elif material == 'glass_panes':
                    material = 'glass_pane'
                elif material == 'iron_ingot':
                    material = 'iron_block'
                try:
                    geometry.placeCuboid(editor,
                                            (x + position['start_x'], z + position['start_y'], y + position['start_z']),
                                            (x + position['end_x'], z + position['end_y'], y + position['end_z']),
                                            Block(material))
                except Exception as e:
                    print(f"exception: {e}")
                    material = 'oak_planks'
                    try:
                        geometry.placeCuboid(editor,
                     (x + position['start_x'], z + position['start_y'], y + position['start_z']),
                         (x + position['end_x'], z + position['end_y'], y + position['end_z']),
                     Block(material))
                    except Exception as e:
                        whether = False
                if not whether:
                    break
                if part['hollow']:
                    try:
                        geometry.placeCuboid(editor, (
                        min(x + position['start_x'] + 1, x + position['end_x']),
                        min(z + position['start_y'] + 1, z + position['end_y']),
                        min(y + position['start_z'] + 1, y + position['end_z'])),
                                         (max(x + position['end_x'] - 1, x + position['start_x']),
                                          max(z + position['end_y'] - 1, z + position['start_y']),
                                          max(y + position['end_z'] - 1, y + position['start_z'])),
                                         Block('air'))
                    except Exception as e:
                        whether = False
                if not whether:
                    break
\end{lstlisting}

\newpage

\subsection{Generating buildings via raw and refined prompts}
\begin{figure}[htbp]
    \centering
\subfigure[Facade (left) and indoor scene (right) of the building generated by GPT-3.5 with the raw prompt in the case of $\neg$ C$\wedge$ $\neg$ S.]{
	\includegraphics[width=0.3\columnwidth, angle=0]{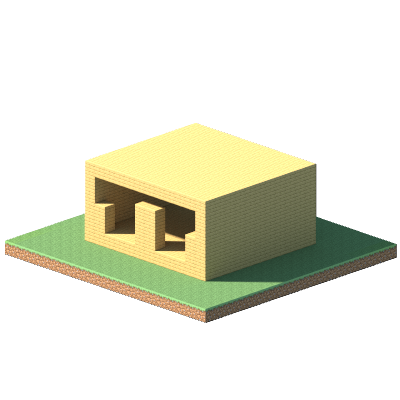}	
\hspace{2em}
 	\includegraphics[width=0.3\columnwidth, angle=0]{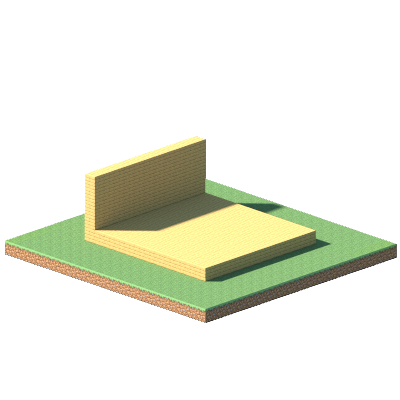}
	\label{fig:3ncns}  
    }\\
    \subfigure[Facade (left) and indoor scene (right) of the building generated by GPT-3.5 with the raw prompt in the case of C$\wedge$ $\neg$ S.]{
	\includegraphics[width=0.3\columnwidth, angle=0]{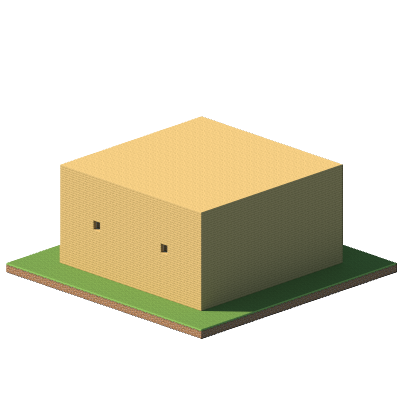}	
\hspace{2em}
 	\includegraphics[width=0.3\columnwidth, angle=0]{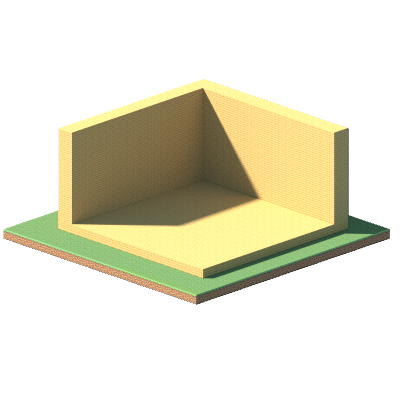}
	\label{fig:3cns}  
    }\\
\subfigure[Facade (left) and indoor scene (right)  of the building generated by GPT-3.5 with the raw prompt in the case of C$\wedge$S.]{
	\includegraphics[width=0.3\columnwidth, angle=0]{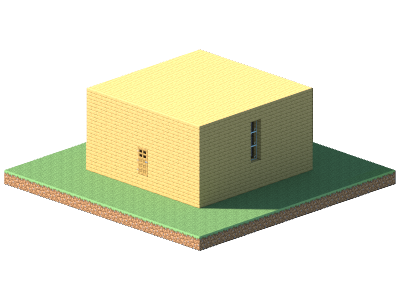}	
\hspace{2em}
 	\includegraphics[width=0.3\columnwidth, angle=0]{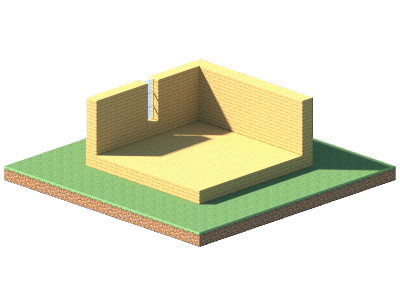}
	\label{fig:3cs}  
    }
\caption{Buildings generated by GPT-3.5 with raw prompts.}
\label{fig:3_raw}
\end{figure}

\begin{figure}[htbp]
    \centering
\subfigure[Building generated by GPT-3.5 with the refined prompt in the case of $\neg$ C$\wedge$ $\neg$ S.]{
\includegraphics[trim=0 50 0 100,width=0.4\columnwidth, angle=0]{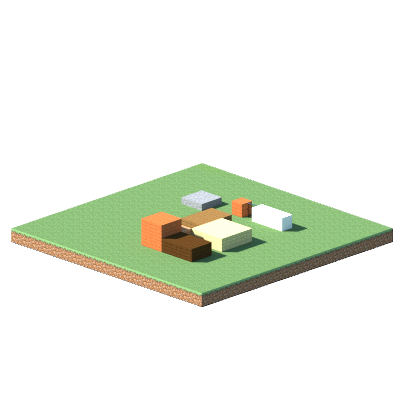}	
	\label{fig:re3ncns}  
    }\hspace{2em}
    \subfigure[Facade scene of building generated by GPT-3.5 with the refined prompt in the case of C$\wedge$ $\neg$ S. The building is solid.]{
\includegraphics[trim=0 50 0 100,width=0.4\columnwidth, angle=0]{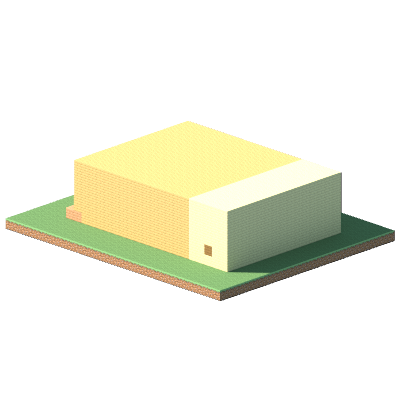}	
	\label{fig:re3cns}  
    }\\
\subfigure[Facade (left) and indoor scene (right) of the building generated by GPT-3.5 with the refined prompt in the case of C$\wedge$S.]{
	\includegraphics[width=0.4\columnwidth, angle=0]{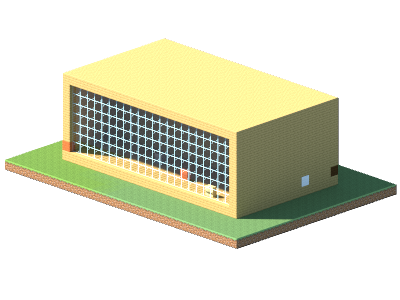}	
\hspace{2em}
 	\includegraphics[width=0.4\columnwidth, angle=0]{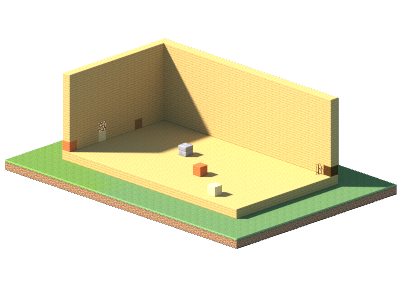}
	\label{fig:re3cs}  
    }
\caption{Buildings generated by GPT-3.5 with refined prompts.}
\label{fig:3_re}
\end{figure}

\begin{figure}[htbp]
    \centering
    
\subfigure[Facade (left) and indoor scene (right) of the building generated by GPT-4 with the raw prompt in the case of $\neg$ C$\wedge$ $\neg$ S.]{
	\includegraphics[trim=0 50 0 100,width=0.4\columnwidth, angle=0]{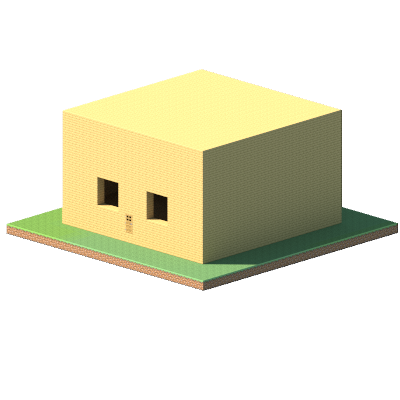}	
\hspace{2em}
 	\includegraphics[trim=0 50 0 100,width=0.4\columnwidth, angle=0]{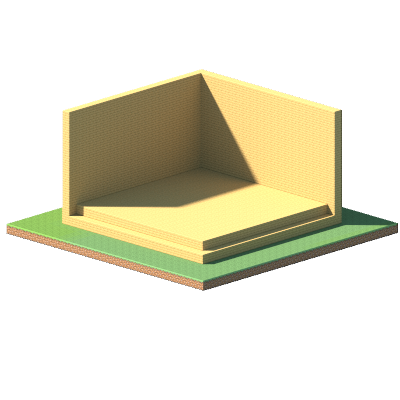}
	\label{fig:4ncns}  
    }\\
    
    \subfigure[Facade (left) and indoor scene (right) of the building generated by GPT-4 with the raw prompt in the case of C$\wedge$ $\neg$ S.]{
	\includegraphics[trim=0 80 0 80,width=0.4\columnwidth, angle=0]{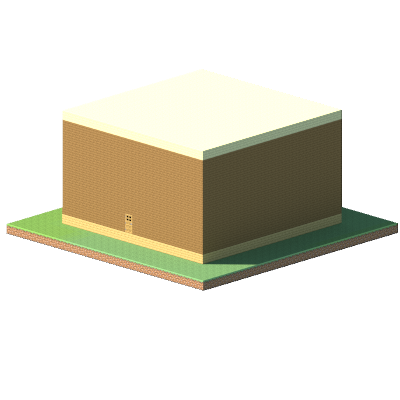}	
\hspace{2em}
 	\includegraphics[trim=0 80 0 80,width=0.4\columnwidth, angle=0]{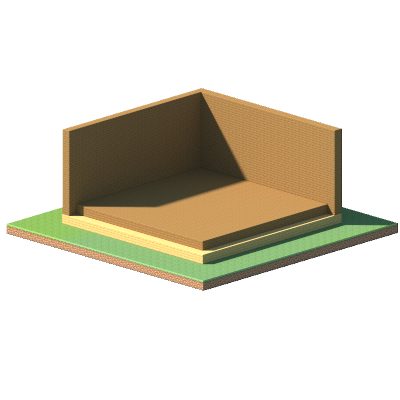}
	\label{fig:4cns}  
    }\\

\subfigure[Facade (left) and indoor scene (right) of the building generated by GPT-4 with the raw prompt in the case of C$\wedge$S.]{
	\includegraphics[width=0.4\columnwidth, angle=0]{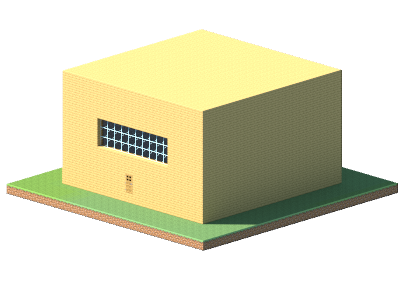}	
\hspace{2em}
 	\includegraphics[width=0.4\columnwidth, angle=0]{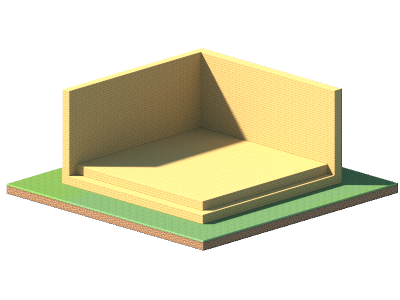}
	\label{fig:4cs}  
    }
\caption{Buildings generated by GPT-4 with raw prompts.}
\label{fig:4_raw}
\end{figure}

\begin{figure}[htbp]
    \centering
\subfigure[Facade (left) and indoor scene (right) of the building generated by GPT-4 with the refined prompt in the case of $\neg$ C$\wedge$ $\neg$ S.]{
	\includegraphics[width=0.4\columnwidth, angle=0]{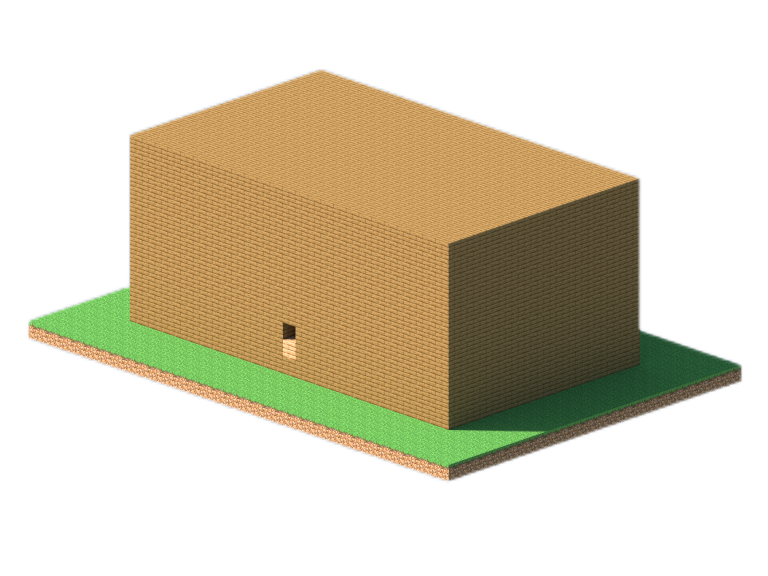}	
\hspace{2em}
 	\includegraphics[width=0.4\columnwidth, angle=0]{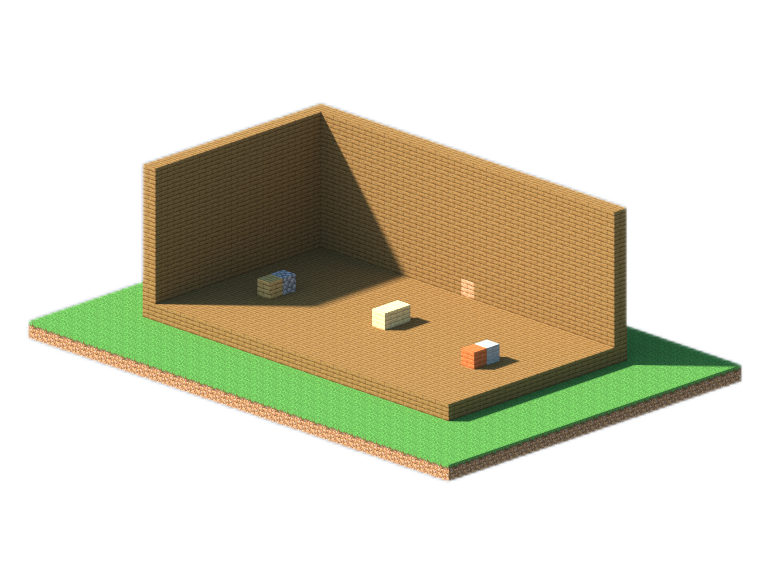}
	\label{fig:re4ncns}  
    }\\
    
    \subfigure[Facade (left) and indoor scene (right) of the building generated by GPT4 with the refined prompt in the case of C$\wedge$ $\neg$ S.]{
	\includegraphics[width=0.4\columnwidth, angle=0]{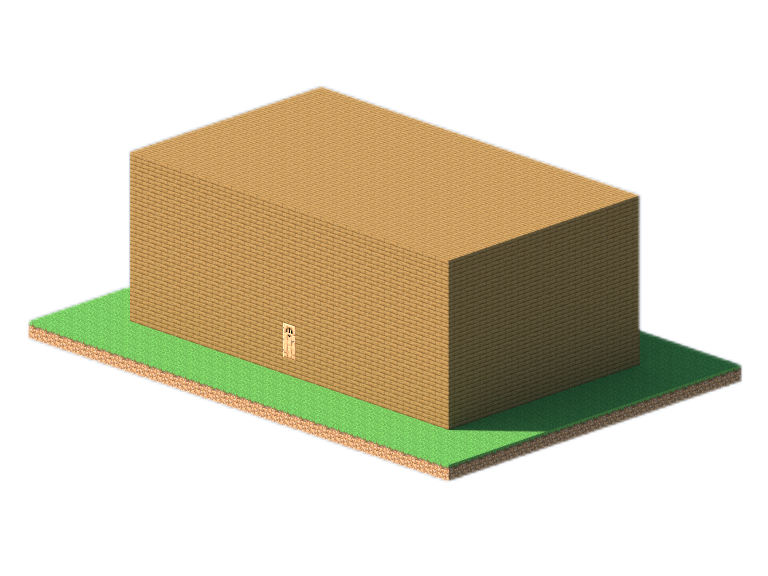}	
\hspace{2em}
 	\includegraphics[width=0.4\columnwidth, angle=0]{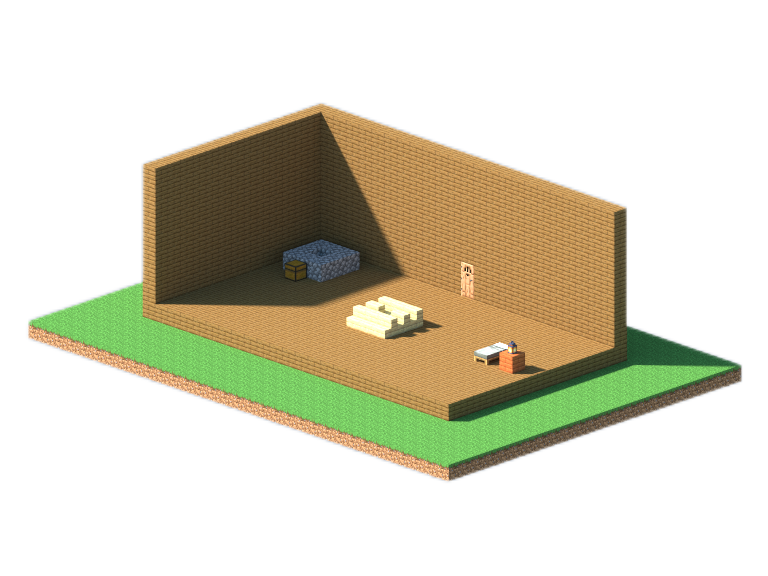}
	\label{fig:re4cns}  
    }\\
    
    \subfigure[Facade (left) and indoor scene (right) of the building generated by GPT4 with the refined prompt in the case of $\neg$ C$\wedge$ S.]{
	\includegraphics[width=0.4\columnwidth, angle=0]{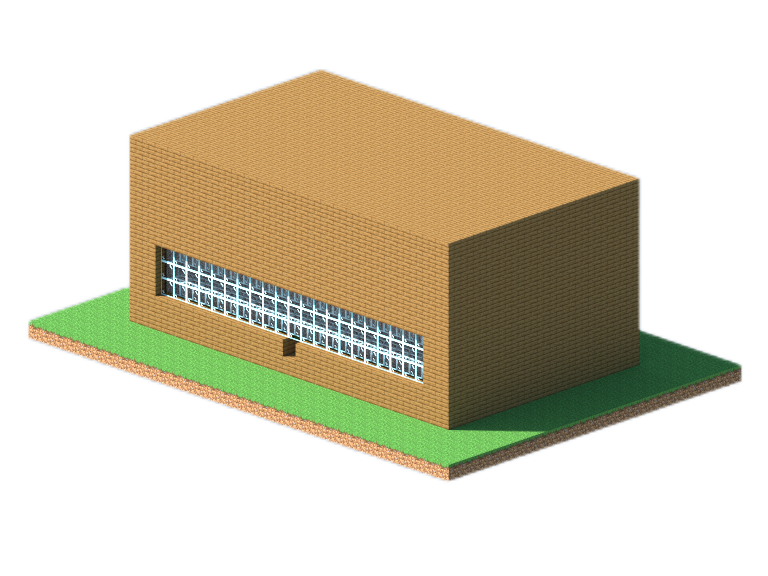}	
\hspace{2em}
 	\includegraphics[width=0.4\columnwidth, angle=0]{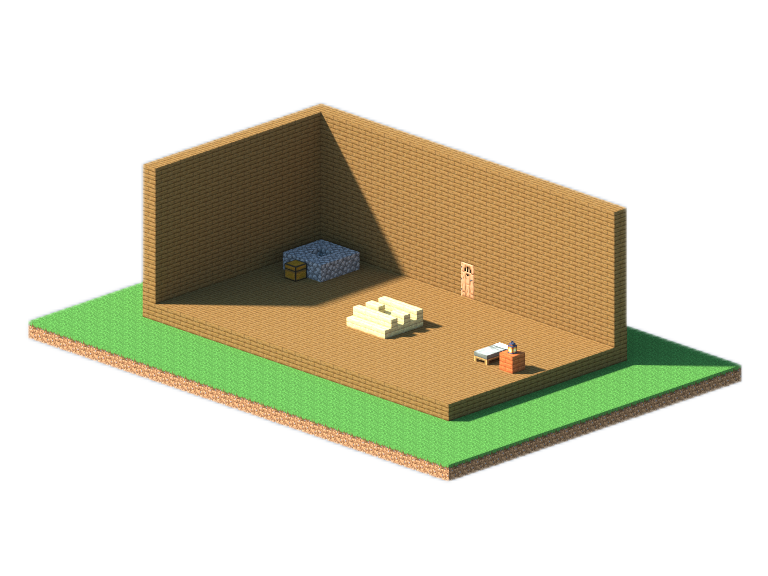}
	\label{fig:re4ncs}  
    }\\

\subfigure[Facade (left) and indoor scene (right) of the building generated by GPT4 with the refined prompt in the case of C$\wedge$S.]{
	\includegraphics[width=0.4\columnwidth, angle=0]{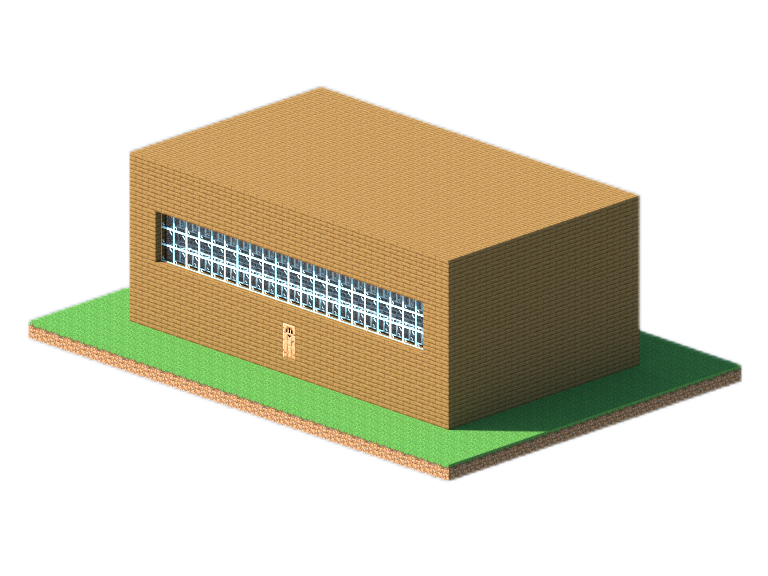}	
\hspace{2em}
 	\includegraphics[width=0.4\columnwidth, angle=0]{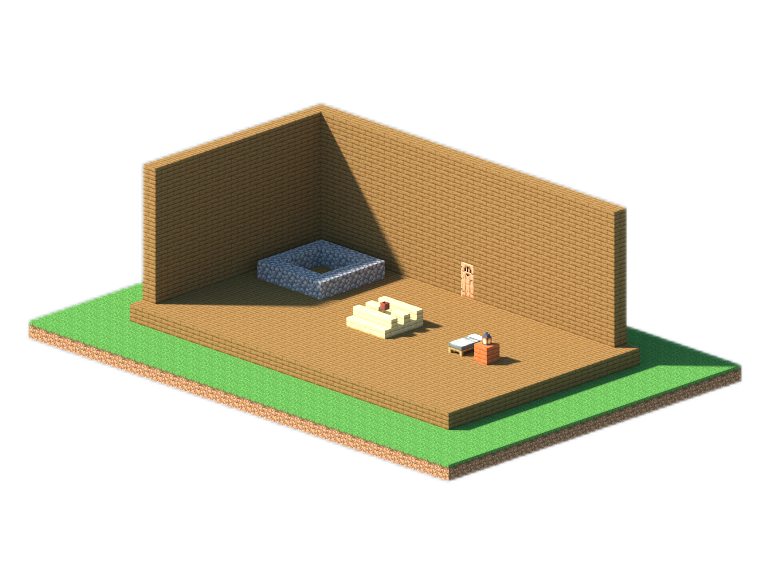}
	\label{fig:re4cs}  
    }
\caption{Buildings generated by GPT-4 with refined prompts.}
\label{fig:4_re}
\end{figure}

This section provides some examples of generated buildings via the prompt ``\textit{A wooden house with windows}" in the case of $\neg$ C$\wedge$ $\neg$ S, C$\wedge$ $\neg$ S, $\neg$ C$\wedge$ S, and C$\wedge$S. Figs.~\ref{fig:3_raw} and \ref{fig:3_re} show buildings generated by GPT-3.5 with raw prompts and refined prompts, respectively. Figs.~\ref{fig:4_raw} and \ref{fig:4_re} show buildings generated by GPT-4 with raw prompts and refined prompts, respectively. 
According to Tab.~\ref{tab:cs}, no building generated by GPT-3.5 with either raw or refined prompts or GPT-4 with raw prompts is in the case of $\neg$ C$\wedge$ S. Notably, there are some buildings that don't have indoor scenes such as Figs.~\ref{fig:re3ncns} and~\ref{fig:re3cns}.

\clearpage

\subsection{Generating buildings via simple prompts}

Four examples of buildings generated via T2BM given simple prompts are presented.

\subsubsection{Generate houses using the prompt ``\textit{A woolen house with a cone roof}"} 
\begin{figure}[htbp]
    \centering
    
\subfigure{
	\includegraphics[trim=0 40 0 200,width=0.4\columnwidth, angle=0]{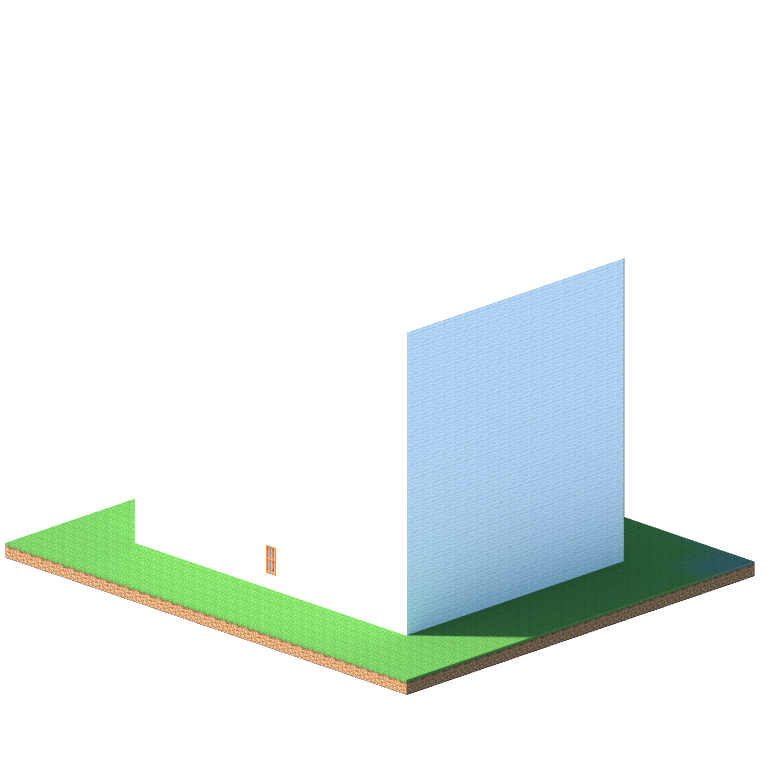}	
\hspace{2em}
 	\includegraphics[trim=0 80 0 170,width=0.4\columnwidth, angle=0]{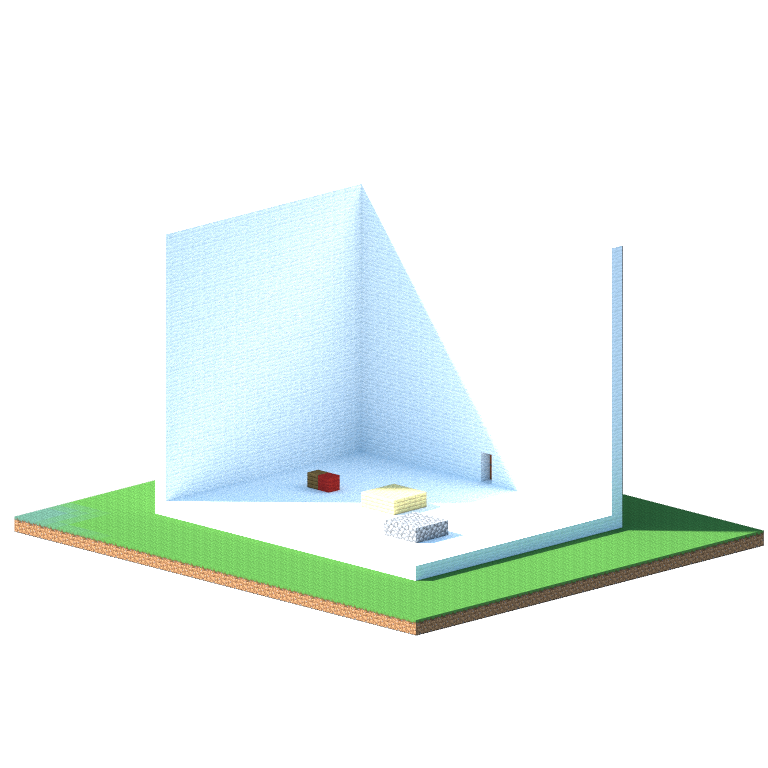}
	\label{fig:woolen}  
    }
\caption{Facade (left) and indoor scene (right) of the house generated by T2BM with prompt ``\textit{A woolen house with a cone roof}".}
\label{fig:house1}
\end{figure}

Fig.~\ref{fig:house1}.

\subsubsection{Generate houses using the prompt ``\textit{A white house with a glass roof}"} 
\begin{figure}[htbp]
    \centering
    
\subfigure{
	\includegraphics[trim=0 150 0 200,width=0.4\columnwidth, angle=0]{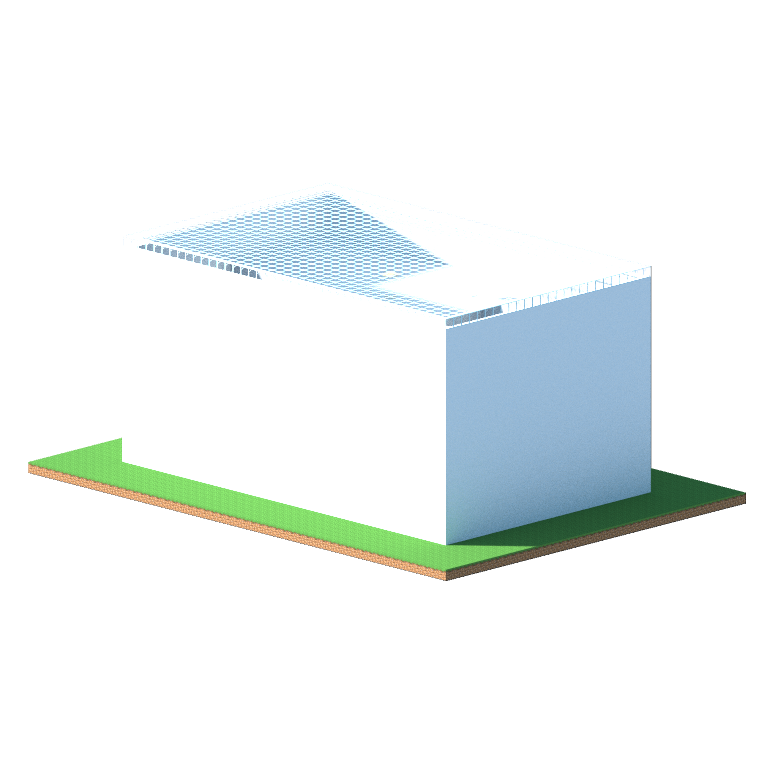}	
\hspace{2em}
 	\includegraphics[trim=0 130 0 220,width=0.4\columnwidth, angle=0]{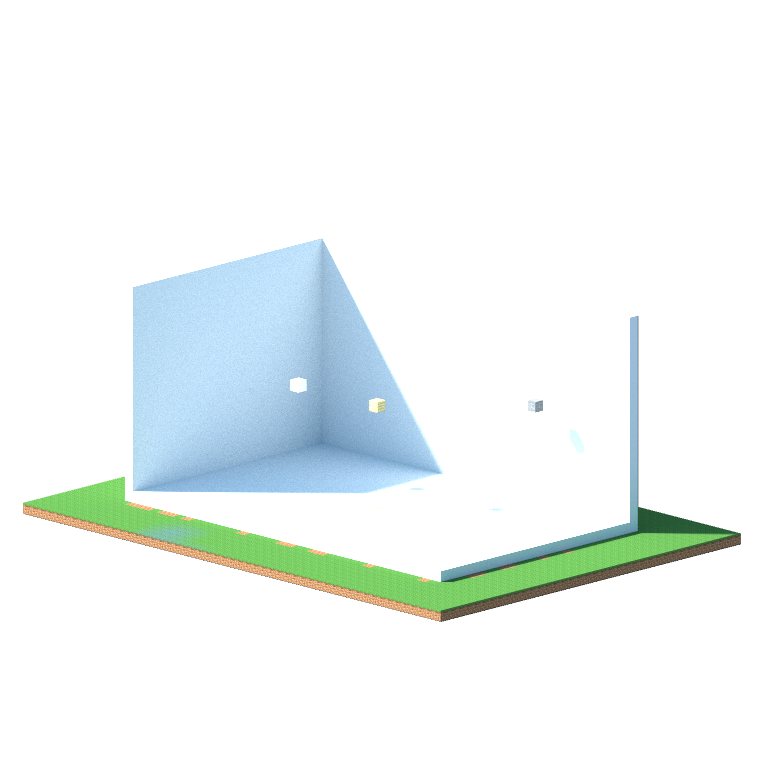}
	\label{fig:white}  
    }
    
\caption{Facade (left) and indoor scene (right) of the house generated by T2BM with prompt ``\textit{A white house with a glass roof}".}
\label{fig:house2}
\end{figure}

Fig.~\ref{fig:house2}.

\subsubsection{Generate houses using the prompt ``\textit{A wooden house with a stone made roof}"} 
\begin{figure}[htbp]
    \centering
    
\subfigure{
	\includegraphics[width=0.4\columnwidth, angle=0]{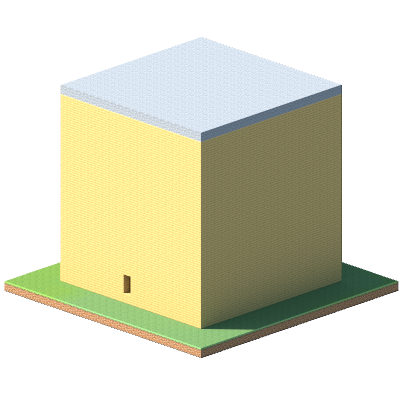}	
\hspace{2em}
 	\includegraphics[width=0.4\columnwidth, angle=0]{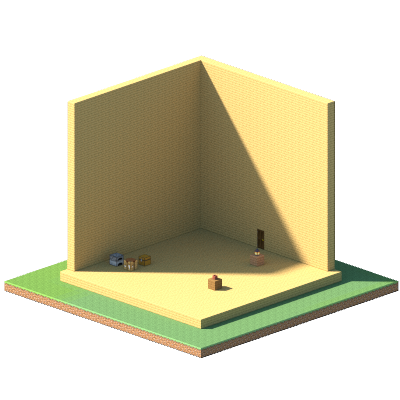}
    }
    
\caption{Facade (left) and indoor scene (right) of the house generated by T2BM with prompt ``\textit{A wooden house with a stone made roof}".}
\label{fig:wooden}
\end{figure}

Fig.~\ref{fig:wooden}.

\subsubsection{Generate houses using the prompt ``\textit{A stone house with a door and a wooden roof}"} 
\begin{figure}[htbp]
    \centering
    
\subfigure{
	\includegraphics[trim=0 50 0 150,width=0.4\columnwidth, angle=0]{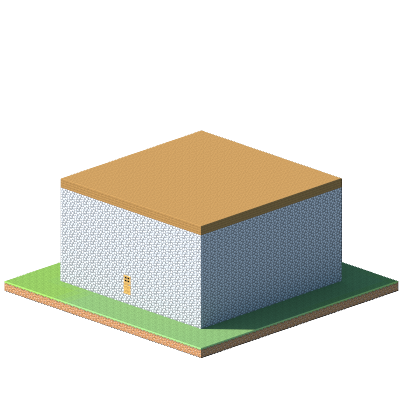}	
\hspace{2em}
 	\includegraphics[trim=0 50 0 150,width=0.4\columnwidth, angle=0]{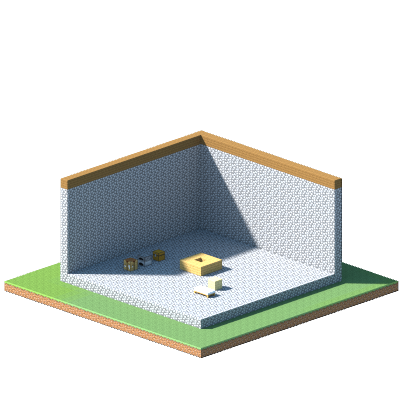}
    }
    
\caption{Facade (left) and indoor scene (right) of the house generated by T2BM with prompt ``\textit{A stone house with a door and a wooden roof}".}
\label{fig:stone}
\end{figure}

Fig.~\ref{fig:stone}.
\subsection{Generating buildings via detailed prompts}

Four examples of generated buildings, including pillbox, snowhouse, nest and tower are given. We directly obtain these descriptions as prompts from \textit{Minecraft Wiki} (\url{https://minecraft.fandom.com/zh/wiki/Minecraft\_Wiki}) and input them to LLMs along with the preset background and format to generate the final building without refining.

\subsubsection{Generating pillboxes}~

\begin{lstlisting}[frame=bt, numbers=none, captionpos=b, abovecaptionskip=5pt, belowcaptionskip=10pt]
Building materials:
1-8 groups of boulders or stones, Torch, Iron gate, glass, Pull rod, button (open door)
Build steps:
A bunker is a single-compartment shelter made entirely of stone,infested_cobblestone, and its walls are usually at least two storeys thick.
A 12x12 size bunker provides players with a bed white_bed, orange_bed, pink_bed, gray_bed, crafting_table, table and table.
Several furnaces or chest Spaces, iron doors must be mounted on different sides of the building (at least on both sides). Stone button stone_button
Or pull lever can be controlled by the inside or outside of the door at the same time. Window glass glass_pane should be installed on each side of the building
\end{lstlisting}

\begin{figure}[htbp]
    \centering
    
\subfigure{
	\includegraphics[trim=0 100 0 150,width=0.35\columnwidth, angle=0]{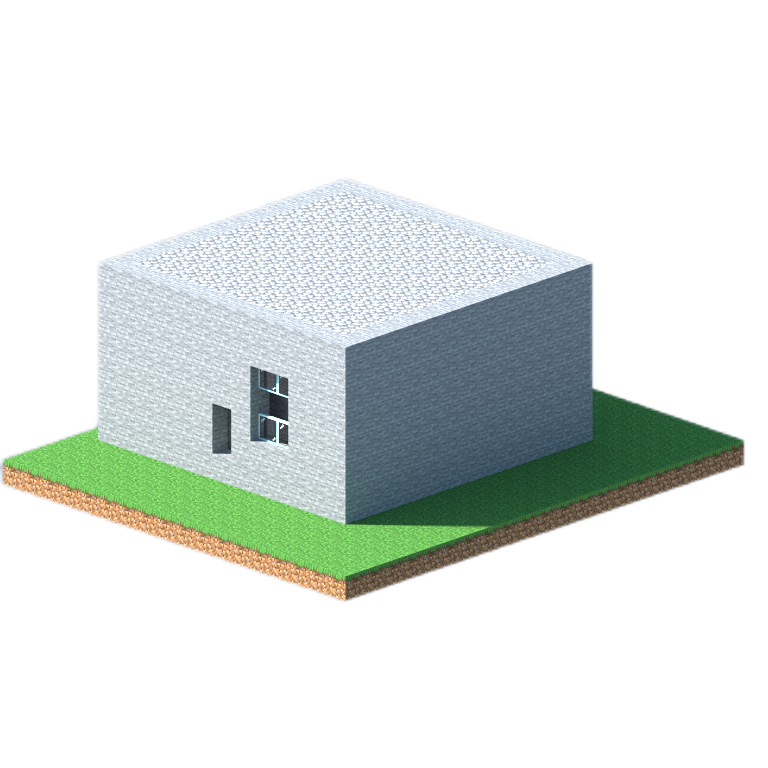}	
\hspace{2em}
 	\includegraphics[trim=0 100 0 150,width=0.35\columnwidth, angle=0]{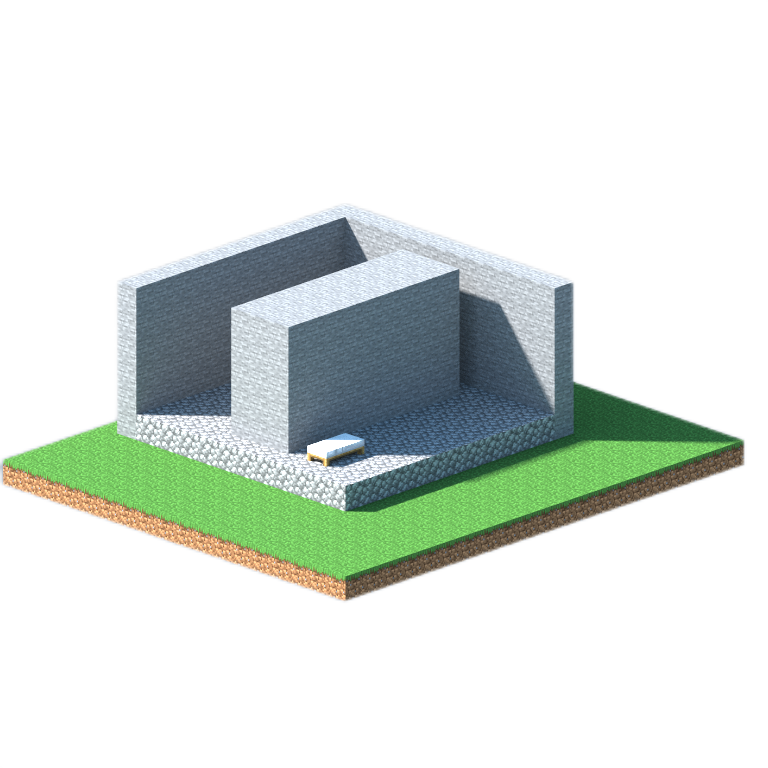}
    }
    
\caption{Facade (left) and indoor scene (right) of a pillbox generated by T2BM.}
\label{fig:pillbox}
\end{figure}

\subsubsection{Generating snowhouses}~
\begin{lstlisting}[frame=bt, numbers=none, captionpos=b, abovecaptionskip=5pt, belowcaptionskip=10pt]
Building materials: 
Lots of snow snow_block
Blue ice, ice blue_ice, ice
Extras materials:
Windows made of ice
A well with liquid water
A furnace (snow blocks do not melt when used, but ice and snow do)
\end{lstlisting}

\begin{figure}[htbp]
    \centering

    \subfigure{
	\includegraphics[trim=0 100 0 200,width=0.35\columnwidth, angle=0]{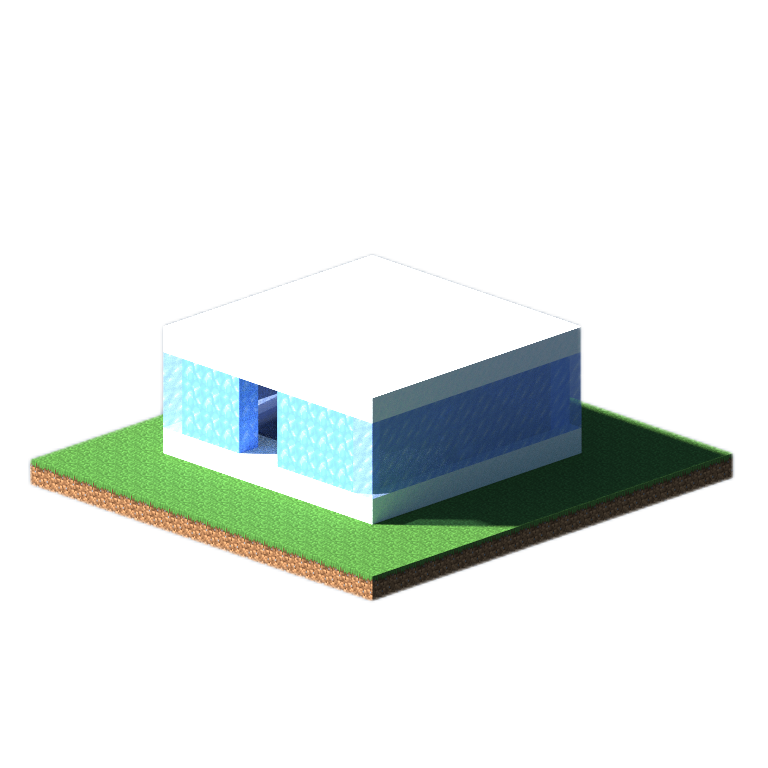}	
\hspace{2em}
 	\includegraphics[trim=0 100 0 200,width=0.35\columnwidth, angle=0]{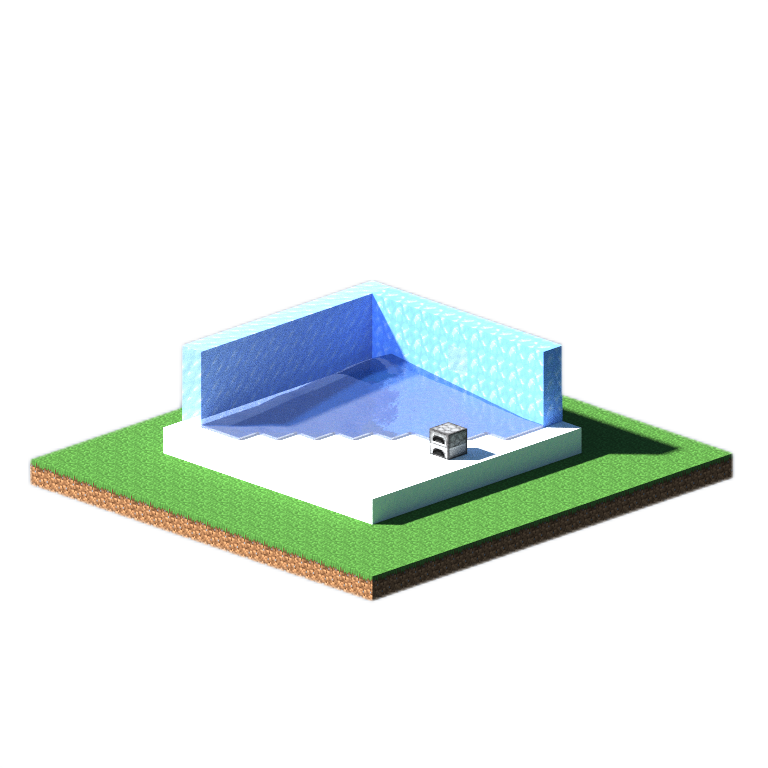}
	\label{fig:snowhouse}  
    }
\caption{Facade (left) and indoor scene (right) of a snowhouse generated by T2BM.}
\label{fig:snow}
\end{figure}

\subsubsection{Generating nests}~

\begin{lstlisting}[frame=bt, numbers=none, captionpos=b, abovecaptionskip=5pt, belowcaptionskip=10pt]
Building materials: 
A lot of dirt or a lot of solid blocks spruce_planks oak_planks.
Description:
A pillar made of blocks that occupies 1 block of space and is 10 blocks high. Then build a platform on it.
Extras materials:
Torches wall_torch
Roof planks
Window glass glass_pane
\end{lstlisting}

\begin{figure}[htbp]
    \centering
\subfigure{
    \includegraphics[trim=0 50 0 100,width=0.35\columnwidth, angle=0]{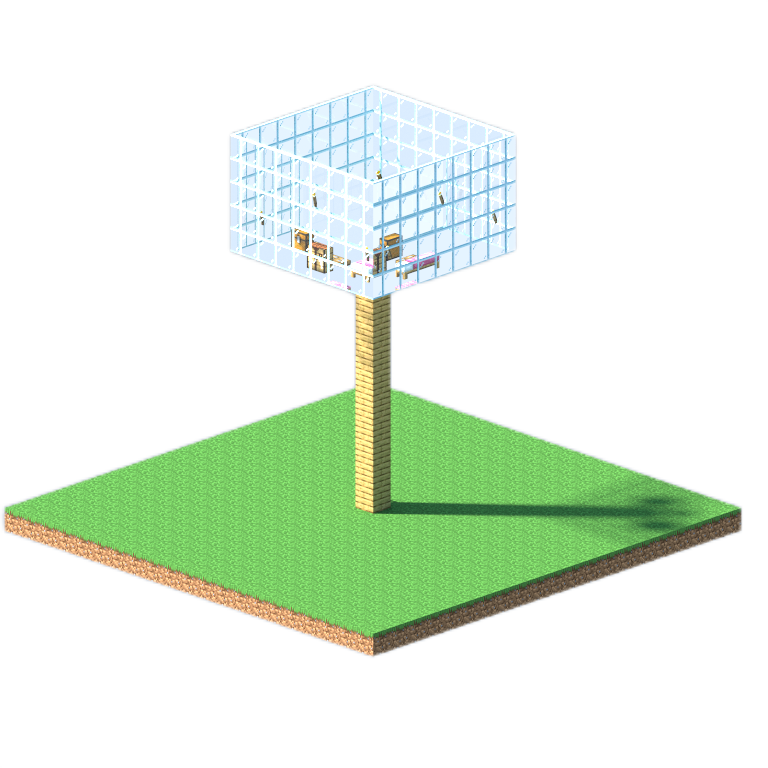}
    \hspace{2em}
 	\includegraphics[trim=0 50 0 100,width=0.35\columnwidth, angle=0]{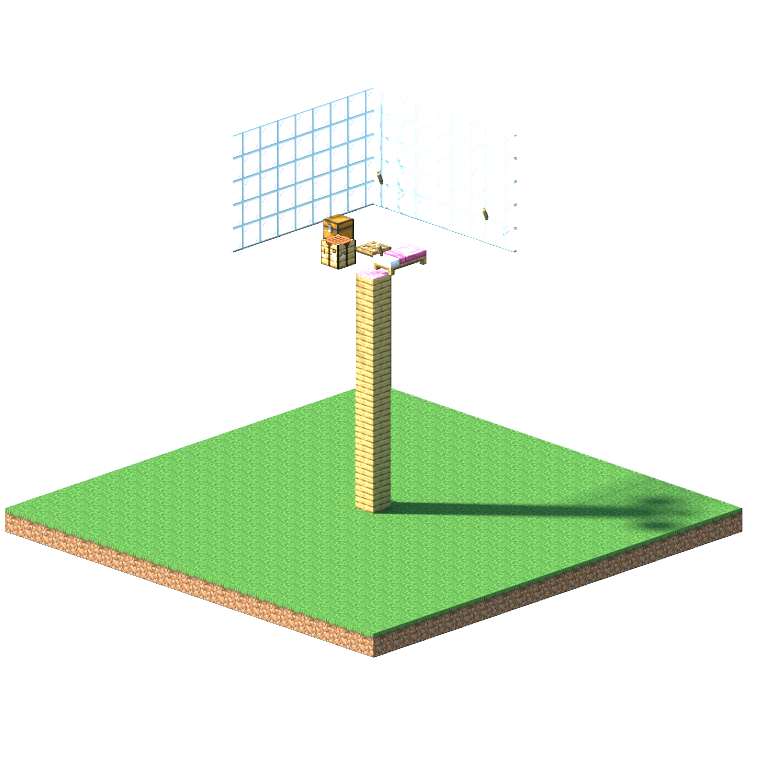}
}

\caption{Facade (left) and indoor scene (right) of a nest generated by T2BM.}
\label{fig:bird}
\end{figure}

\subsubsection{Generating towers}~

\begin{lstlisting}[frame=bt, numbers=none, captionpos=b, abovecaptionskip=5pt, belowcaptionskip=10pt]
Building materials: 
stone.
Build steps:
Build a 3x3x2 tower.
Add a block to each edge of the tower, then add another block to the outside of those blocks, removing the previous block so that the later block is half-overhang. On top of these blocks, build a 5x5 frame.
Plus torches wall_torch.
Dig a 2x2x1 hole in the center of the tower
\end{lstlisting}

\begin{figure}[htbp]
    \centering
 \includegraphics[trim=0 50 0 100,width=.4\columnwidth, angle=0]{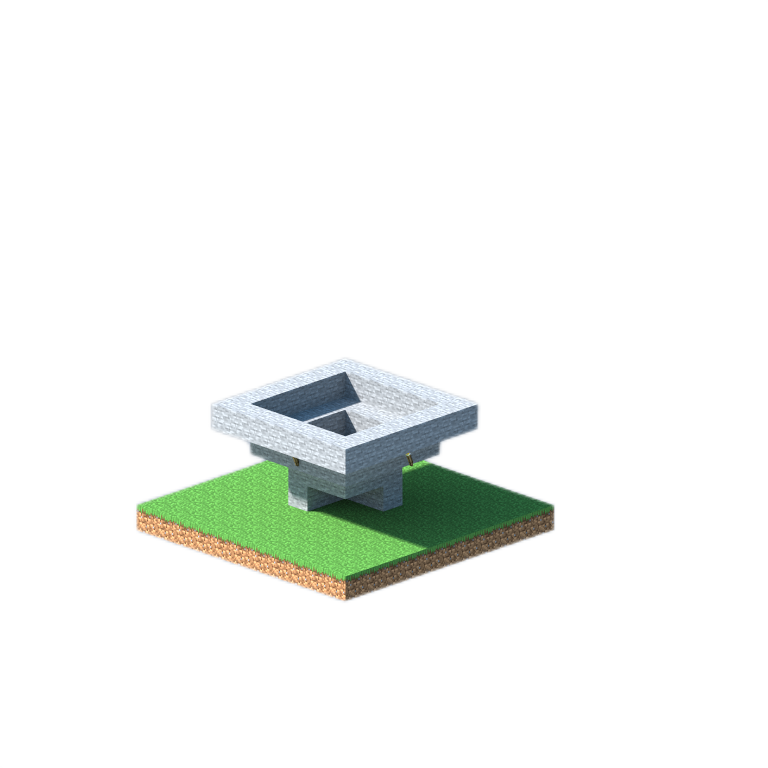}	
\caption{Facade scene of a tower generated by T2BM.}
\label{fig:lookout}
\end{figure}

\end{document}